\documentclass[10pt,journal,compsoc]{IEEEtran}
\usepackage{amsmath,amssymb,amsfonts}
\usepackage{algorithmic}
\usepackage{graphicx}
\usepackage{textcomp}
\newlength{\xfigwd}
\usepackage{multirow}
\usepackage{url}
\usepackage{threeparttable}
\usepackage{footmisc}
\usepackage{color}
\usepackage{diagbox}

\newcommand{\Tref}[1]{Table~\ref{#1}}
\newcommand{\Eref}[1]{Equation~(\ref{#1})}
\newcommand{\Fref}[1]{Figure~\ref{#1}}
\newcommand{\Sref}[1]{Section~\ref{#1}}

\ifCLASSOPTIONcompsoc
  \usepackage[nocompress]{cite}
\else
  \usepackage{cite}
\fi
\ifCLASSINFOpdf
\else
\fi
\begin{document}
%
\title{Efficient Semantic Image Synthesis via Class-Adaptive Normalization}
%
%
%
%

\author{Zhentao Tan$^*$, Dongdong Chen$^*$, Qi Chu$^\dagger$, Menglei Chai, Jing Liao, Mingming He,\\ Lu Yuan, Gang Hua, ~\IEEEmembership{Fellow,~IEEE}, Nenghai Yu
    \IEEEcompsocitemizethanks{
			\IEEEcompsocthanksitem Zhentao Tan, Qi Chu and Nenghai Yu are with School of Information Science and Technology, University of Science and Technology of China; (Email: tzt@mail.ustc.edu.cn; (qchu, ynh)@ustc.edu.cn).
            \IEEEcompsocthanksitem Dongdong Chen and Lu Yuan are with Microsoft Cloud AI (cddlyf@gmail.com, luyuan@microsoft.com).
            \IEEEcompsocthanksitem Menglei Chai is with Snap Inc. (cmlatsim@gmail.com).
            \IEEEcompsocthanksitem Jing Liao is with City University of Hong Kong (jingliao@cityu.edu.hk).
            \IEEEcompsocthanksitem Mingming He is with USC Institute for Creative Technologies (hmm.lillian@gmail.com).
            \IEEEcompsocthanksitem Gang Hua is with Wormpex AI Research LLC (ganghua@gmail.com).
            \IEEEcompsocthanksitem $*$ Zhentao Tan and Dongdong Chen are co-first authors.
            \IEEEcompsocthanksitem $\dagger$ Qi chu is the corresponding author.
    }
}
%
%

\markboth{IEEE TRANSACTIONS ON PATTERN ANALYSIS AND MACHINE INTELLIGENCE}%
{Tan \MakeLowercase{\textit{et al.}}: Efficient Semantic Image Synthesis via Class-Adaptive Normalization}
%



\IEEEtitleabstractindextext{%
\begin{abstract}
Spatially-adaptive normalization (SPADE) is remarkably successful recently in conditional semantic image synthesis~\cite{park2019semantic}, which modulates the normalized activation with spatially-varying transformations learned from semantic layouts, to prevent the semantic information from being washed away. Despite its impressive performance, a more thorough understanding of the advantages inside the box is still highly demanded to help reduce the significant computation and parameter overhead introduced by this novel structure. In this paper, from a return-on-investment point of view, we conduct an in-depth analysis of the effectiveness of this spatially-adaptive normalization and observe that its modulation parameters benefit more from semantic-awareness rather than spatial-adaptiveness, especially for high-resolution input masks.
Inspired by this observation, we propose class-adaptive normalization (CLADE), a lightweight but equally-effective variant that is only adaptive to semantic class. In order to further improve spatial-adaptiveness, we introduce intra-class positional map encoding calculated from semantic layouts to modulate the normalization parameters of CLADE and propose a truly spatially-adaptive variant of CLADE, namely CLADE-ICPE. 
Through extensive experiments on multiple challenging datasets, we demonstrate that the proposed CLADE can be generalized to different SPADE-based methods while achieving comparable generation quality compared to SPADE, but it is much more efficient with fewer extra parameters and lower computational cost. The code and pretrained models are available at \url{https://github.com/tzt101/CLADE.git}.
\end{abstract}

}

\maketitle

\IEEEdisplaynontitleabstractindextext

%
\IEEEpeerreviewmaketitle

\IEEEraisesectionheading{\section{Introduction}}
\label{sec:introduction}
\IEEEPARstart{I}{mage} synthesis has made great progress recently thanks to the advances of deep generative models. The latest successes, such as StyleGAN~\cite{karras2019style,karras2020analyzing}, are already capable of producing highly realistic images from random latent codes. Yet conditional image synthesis, the task of generating photo-realistic images conditioned on some input data, is still very challenging. In this work, we focus on semantic image synthesis, a specific conditional image generation task that aims at converting a semantic segmentation mask into a photo-realistic image.

To tackle this problem, some previous methods~\cite{isola2017image,wang2018high} directly feed the semantic segmentation mask to the conventional deep network architecture built by stacking convolution, normalization, and nonlinearity layers. However, as pointed out in~\cite{park2019semantic}, common normalization layers like instance normalization~\cite{ulyanov2016instance} tend to wash away the semantic information, especially for flat segmentation masks. To compensate for the information loss, a novel spatially-adaptive normalization, SPADE~\cite{park2019semantic}, is proposed, which modulates the normalized activation in a spatially-adaptive manner, conditioned on the input segmentation mask. Therefore, by replacing all the common normalization layers with SPADE blocks, the semantic information can be successfully propagated throughout the network, which can improve performance in terms of visual fidelity and spatial alignment.

\begin{figure*}[tp]
\begin{center}
\includegraphics[width=1\linewidth]{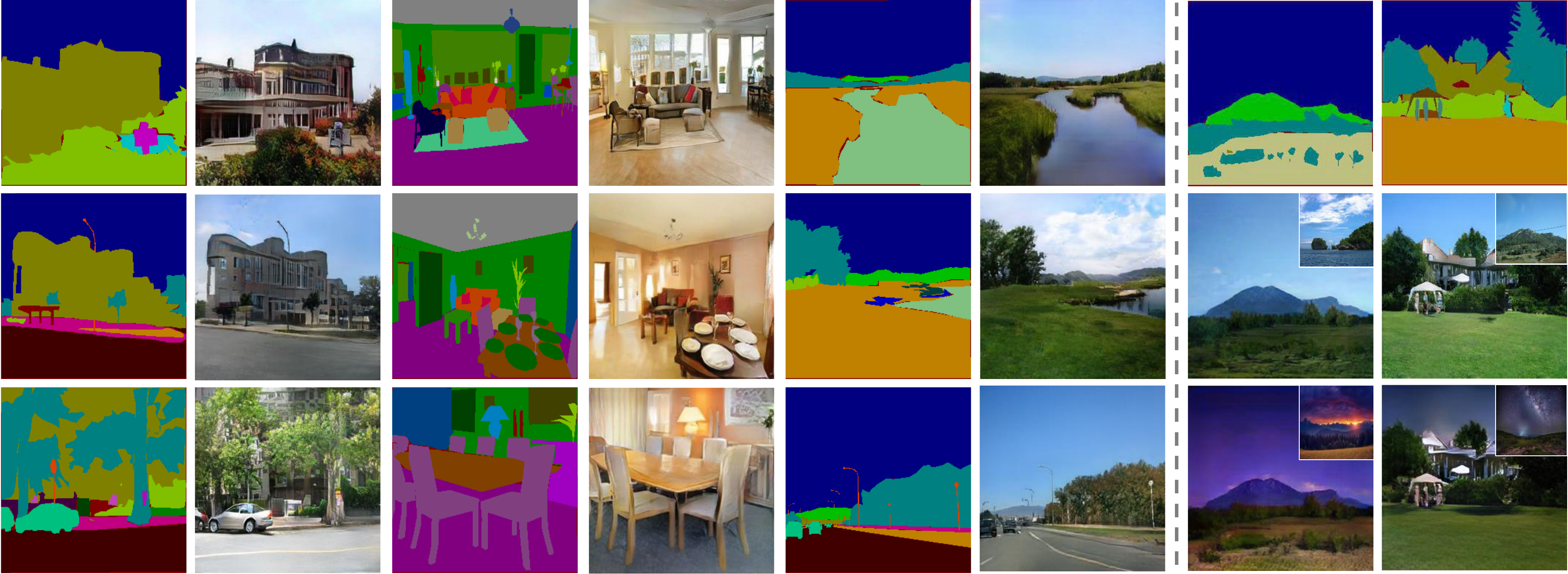}
\caption{Some semantic image synthesis results produced by our method.
Our method can not only handle the synthesis from a pure semantic segmentation mask (left six columns) but also support controllable synthesis via different reference style images (right two columns). For the controllable image generation, the input semantic masks are given on the first row, and the reference style images are displayed in the upper right corner of the generated results (second and third rows).
}
\label{fig:teaser}
\end{center}
\end{figure*}

Despite the effectiveness of the spatially-adaptive normalization, it introduces significant memory and computation overhead, which limits its applications as a general normalization in other models. In addition, its advantages have not been fully uncovered yet. Is \textit{spatial-adaptiveness} the sole or main reason for its superior performance? Does there exist any better design that can improve efficiency without compromising the resulting quality? In this paper, we try to answer these questions by analyzing it in depth. Our key observation is that \textit{semantic-awareness} may actually contribute much more than the \textit{spatial-adaptiveness}. In fact, since the two-layer modulation network used to regress the transformation parameters is so shallow, the resulting denormalization parameters are almost spatial-invariant within regions with the same semantic class, especially for high-resolution input masks. Meanwhile, given that a SPADE block is placed before almost every convolutional layer, such redundancy is recurring multiple times in the generation pass, which can easily lead to a heavy amount of unnecessary computation and parameter overhead. 

Motivated by this observation, we propose a novel normalization layer, namely CLass-Adaptive (DE)normalization (CLADE). Different from the spatially adaptive solution of SPADE, CLADE instead uses the input semantic mask to modulate the normalized activation in a \emph{class-adaptive} manner. Specifically, CLADE is only adaptive to different semantic classes to maintain the crucial semantic-awareness property, independent of the spatial position, semantic shape, or layout of the semantic mask. Thanks to this lightweight design, CLADE is surprisingly simple to implement and requires no extra modulation network. Therefore, its computation and parameter overhead is almost negligible compared with SPADE, making it a better alternative to those conventional normalization layers. Take the generator for the \textit{ADE20k} dataset~\cite{zhou2017scene} as an example, the extra parameter and computation cost introduced by CLADE is only $4.57\%$ and $0.07\%$ while that of SPADE is $39.21\%$ and $234.73\%$ respectively.

Although \emph{class-adaptiveness} greatly reduces the computational overhead and achieves excellent performance, we believe that spatial-adaptiveness could still be beneficial to better semantic synthesis. To enhance the spatial-adaptiveness expected by SPADE, we further propose to utilize an extra positional encoding map representing the intra-class spatial variance, which defines the normalized relative distance from each pixel to its semantic object center. This positional encoding is then integrated into the CLADE modulation parameters and makes them spatially-adaptive in the regions with the same semantic class. This can be viewed as a spatially-adaptive variant of CLADE, namely CLADE-ICPE.

To demonstrate the effectiveness and efficiency of CLADE, we conduct extensive experiments on multiple challenging datasets, including \textit{Cityscapes}~\cite{cordts2016cityscapes}, \textit{COCO-Stuff}~\cite{caesar2018coco}, and \textit{ADE20k} (including \textit{ADE20k-outdoor})~\cite{zhou2017scene}. Without bells and whistles, just by replacing all the SPADE layers with CLADE, comparable performance can be achieved with much smaller model size and much lower computation cost.
Some visual results are given in \Fref{fig:teaser}.

\section{Related Works}

\subsection{Generative Adversarial Networks}
In recent years, image synthesis has achieved significant progress thanks to the emergence of generative adversarial networks (GANs)~\cite{goodfellow2014generative}. This adversarial training strategy enables the generator network to synthesize images with semantic meaning from a random noise. Starting from the early work~\cite{goodfellow2014generative}, many following works have been proposed from different aspects. For example, to make the network training more stable, some works~\cite{arjovsky2017wasserstein,salimans2016improved,mao2017least} propose improvements based on the loss functions. DCGAN~\cite{radford2015unsupervised} proposes a set of constraints on the architectural topology of Convolutional GANs that make them stable to train in most settings. The work in~\cite{qin2020does} shows that the effectiveness of many GAN loss functions actually comes from the Lipschitzness of the discriminator network.
For higher resolution and quality, ProgressiveGAN~\cite{karras2017progressive} designs a training strategy to gradually synthesize high-resolution images. BigGAN~\cite{brock2018large} proposes to train the network on a large-scale image dataset to improve the capabilities of generator. The recent works~\cite{karras2019style,karras2020analyzing} have not only pursued the realistic image synthesis, but also attempted to improve the accurate control of the synthesized image through the exploration of latent code. Different from this work, we are more interested in controlling the synthesized image in a more intuitive way, by using additional conditional inputs to control the synthesis results.

\subsection{Conditional Image Synthesis}
Instead of generation from a random noise, conditional image synthesis refers to the task of generating photo-realistic images conditioned on the input such as 
texts~\cite{hong2018inferring,reed2016generative,xu2018attngan,zhang2017stackgan} and images~\cite{huang2018multimodal,isola2017image,liu2017unsupervised,zhu2017unpaired,oza2019semi,park2019semantic}. Our work focuses on a special form of conditional image synthesis that aims at generating photo-realistic images conditioned on input segmentation masks, which is called semantic image synthesis. 

For this task, many impressive works have been proposed in the past several years. One of the most representative works is pix2pix~\cite{isola2017image}, which proposes a unified image-to-image translation framework based on the conditional generative adversarial network. To further improve its quality or enable more functionality, many following works have appeared, such as pix2pixHD~\cite{wang2018high}, SIMS~\cite{qi2018semi}, and SPADE~\cite{park2019semantic}. SPADE proposes a spatial-varying normalization layer for the first time and has a profound impact as a basic backbone. Many recent works for different downstream tasks have used this architecture, such as semantic image synthesis~\cite{dundar2020panoptic,zhu2020semantically,zheng2019example}, portrait synthesis or editing~\cite{zhu2020sean,tan2020michigan} and semantic view synthesis~\cite{huang2020semantic}. Other works~\cite{jiang2020tsit,zhang2020cross,tan2021diverse}, although not using SPADE directly, are inspired by it to introduce spatial-adaptiveness or approximate spatial-adaptiveness into normalization layers. Despite the success of SPADE, its efficiency is often neglected and understudied in the community. In this paper, we conduct an in-depth analysis of its superiority and propose a new efficient and effective normalization layer.  

\subsection{Normalization Layers}
In the deep learning era, normalization layers play a vital role in achieving better convergence and performance, especially for deep networks. They follow a similar operating logic, which first normalizes the input features into zero mean and unit deviation, and then modulates the normalized features with learnable modulation scale/shift parameters.

Existing normalization layers can be generally divided into two different types: unconditional and conditional. Typical unconditional normalization layers include Batch Normalization (BN)~\cite{ioffe2015batch}, Instance Normalization (IN)~\cite{ulyanov2016instance}, Group Normalization (GN)~\cite{wu2018group} and Positional Normalization (PONO)~\cite{li2019positional}. Compared to unconditional normalization, the behavior of conditional normalization is not static and depends on the external input. Conditional Instance Normalization (Conditional IN)~\cite{dumoulin2016learned} and Adaptive Instance Normalization (AdaIN)~\cite{huang2017arbitrary} are two popular conditional normalization layers originally designed for style transfer. To transfer the style from one image to another, they model the style information into the modulation scale/shift parameters. 

For semantic image synthesis, most previous works~\cite{park2019semantic} just leveraged unconditional normalization layers BN or IN in their networks. Recently, Park \emph{et al.}~\cite{park2019semantic} point out that common normalization layers used in the existing methods tend to ``wash away'' semantic information when applied to flat segmentation masks. To compensate for the missing information, they innovatively propose a new spatially-adaptive normalization layer named SPADE. Different from common normalization layers, SPADE puts the semantic information back by making the modulation parameters be the function of semantic mask in a spatially-adaptive way.
Based on our analysis and observation that the semantic-awareness is the possible essential property leading to the superior performance of SPADE rather than the spatially-adaptiveness, we propose CLADE, a normalization layer that can achieve comparable performance as SPADE but with negligible cost.

\section{Semantic Image Synthesis}
Conditioned on a semantic segmentation map $m\in\mathbb{L}^{H\times W}$, semantic image synthesis aims at generating a corresponding high-quality realistic image $I$. Here, $\mathbb{L}$ is the set of class integers that denote different semantic categories. $H$ and $W$ are the target image height and width.

Most vanilla synthesis networks, like pix2pix~\cite{isola2017image} and pix2pixHD~\cite{wang2018high}, adopt a similar network structure concatenating repeated blocks of convolutional, normalization and nonlinearity layers. Among them, normalization layers are essential for better convergence and performance. They can be generally formulated as:
\begin{equation}
    \hat{x}^{in}_{i,j,k} = \frac{x^{in}_{i,j,k}-\mu_{i,j,k}}{\sigma_{i,j,k}},\quad
    x^{out}_{i,j,k} = \gamma_{i,j,k} \hat{x}^{in}_{i,j,k} + \beta_{i,j,k},
\end{equation}
with the indices of width, height, channel denoted as $i,j,k$. In what follows, for the simplicity of notation, these subscripts will be omitted if the variable is independent of them. Specifically, the input feature $x^{in}$ is first normalized with the mean $\mu$ and standard deviation $\sigma$ (normalization step), and then modulated with the learned scale $\gamma$ and shift $\beta$ (modulation step). For most common normalization layers such as BN~\cite{isola2017image} and IN~\cite{ulyanov2016instance}, all four parameters are calculated in a channel-wise manner (independent of $i,j$), with the modulating parameters $\gamma$ and $\beta$ independent of $x^{in}$.

\subsection{Revisit Spatially-Adaptive Normalization}
\begin{figure*}[tp]
  \centering
  \includegraphics[width=\linewidth]{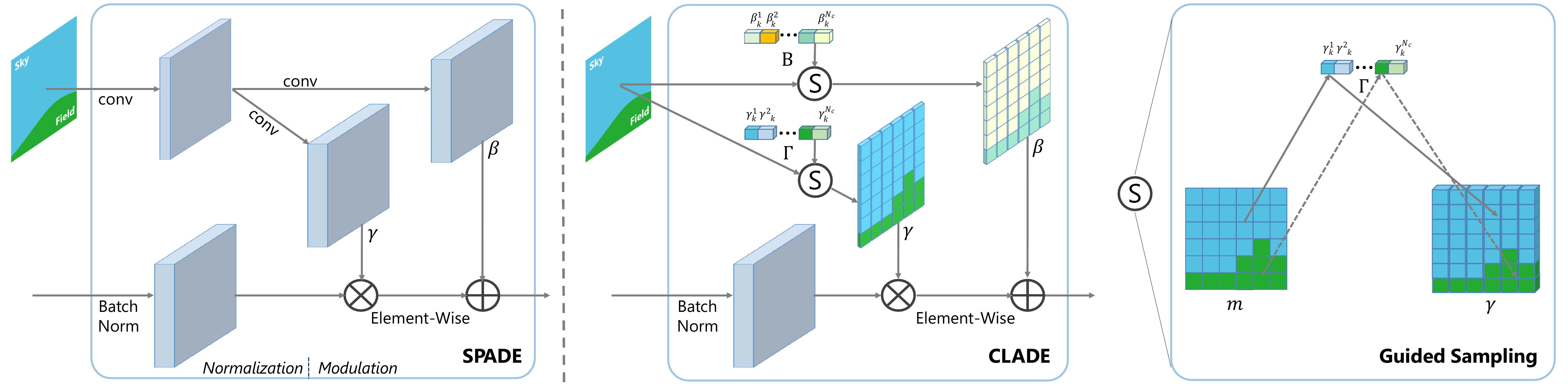}
  \caption{The illustration diagrams of SPADE (left) and our class-adaptive normalization layer CLADE with a guided sampling operation (right). Using a shallow modulation network consisting of two convolutional layers to model the modulation parameters $\gamma,\beta$ as the function of input semantic mask, SPADE can add the semantic information lost in the normalization step back. Unlike SPADE, CLADE does not introduce any external modulation network but instead uses an efficient guided sampling operation to sample class-adaptive modulation parameters for each semantic region.
  }
  \label{fig:SPADE-CLADE}
\end{figure*}

As pointed out in~\cite{park2019semantic}, one common issue of the aforementioned normalization layers is that they tend to wash away the semantic information on flat segmentation masks $m$ in image synthesis.
Motivated by this observation, a new spatially-adaptive normalization layer, namely SPADE, is proposed in~\cite{park2019semantic}. By making the modulation parameters $\gamma$ and $\beta$ be functions of the input mask $m$, \textit{i.e.}, $\gamma_{i,j,k}(m)$ and $\beta_{i,j,k}(m)$, the semantic information, which is lost after the normalization step, will be added back during the modulation step. The functions of $\gamma_{i,j,k}(\cdot)$ and $\beta_{i,j,k}(\cdot)$ are both implemented with a shallow modulation network consisting of two convolutional layers, as illustrated in the left of \Fref{fig:SPADE-CLADE}. By replacing all the normalization layers with SPADE, the generation network proposed in~\cite{park2019semantic} can achieve much better synthesis results than previous methods like pix2pixHD~\cite{wang2018high}.

\begin{figure}[tp]
  \centering
  \includegraphics[width=0.99\linewidth]{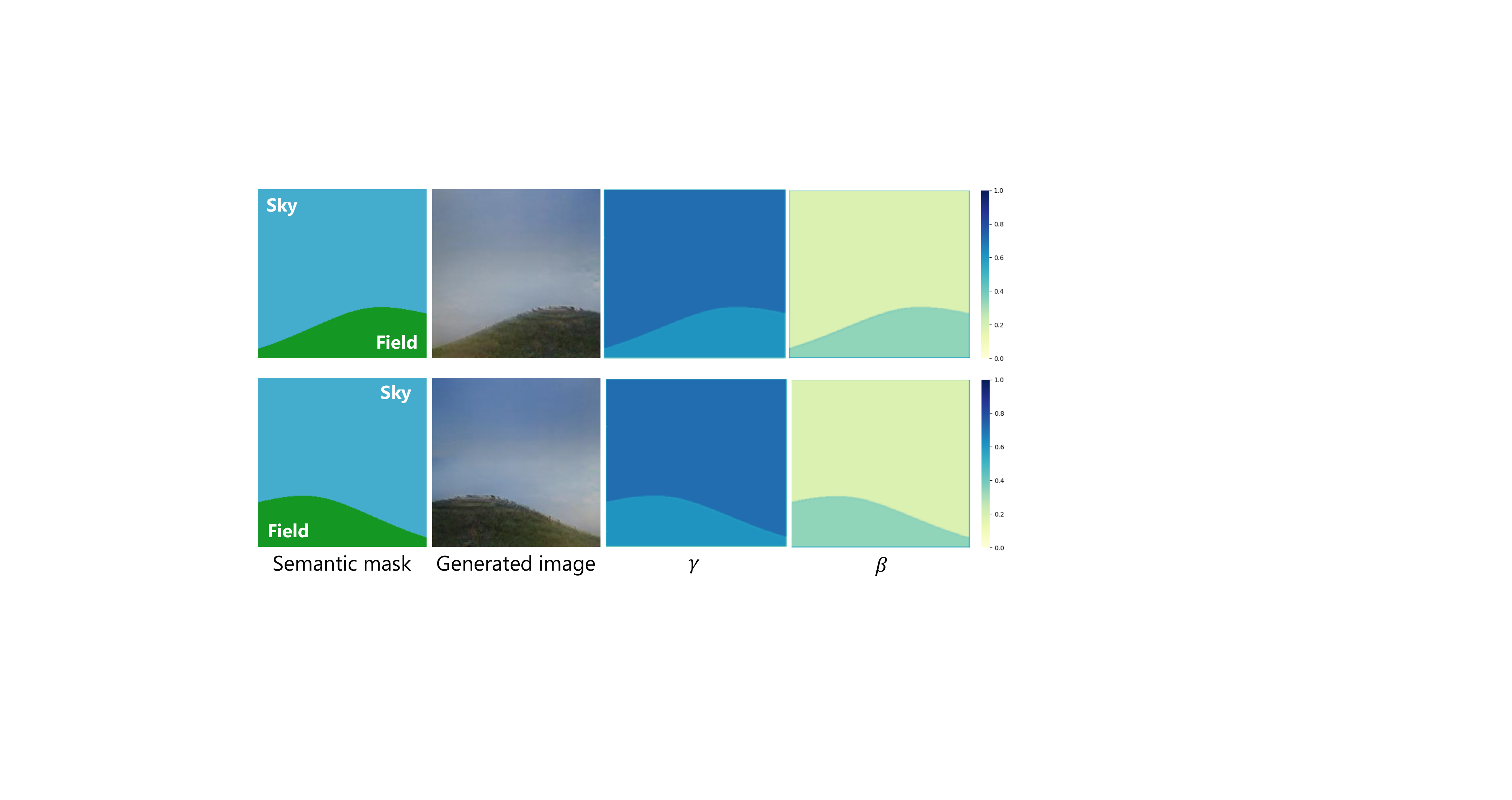}
  \caption{Visualization of learned modulation parameters $\gamma,\beta$ at the shallowest layer for two example semantic masks from the \textit{ADE20k} dataset, where the original pre-trained SPADE generator is used. Obviously, $\gamma,\beta$ for the same semantic class are almost identical within each semantic region. }
  \label{fig:feature}
\end{figure}

\begin{figure}[tp]
  \centering
  \includegraphics[width=0.99\linewidth]{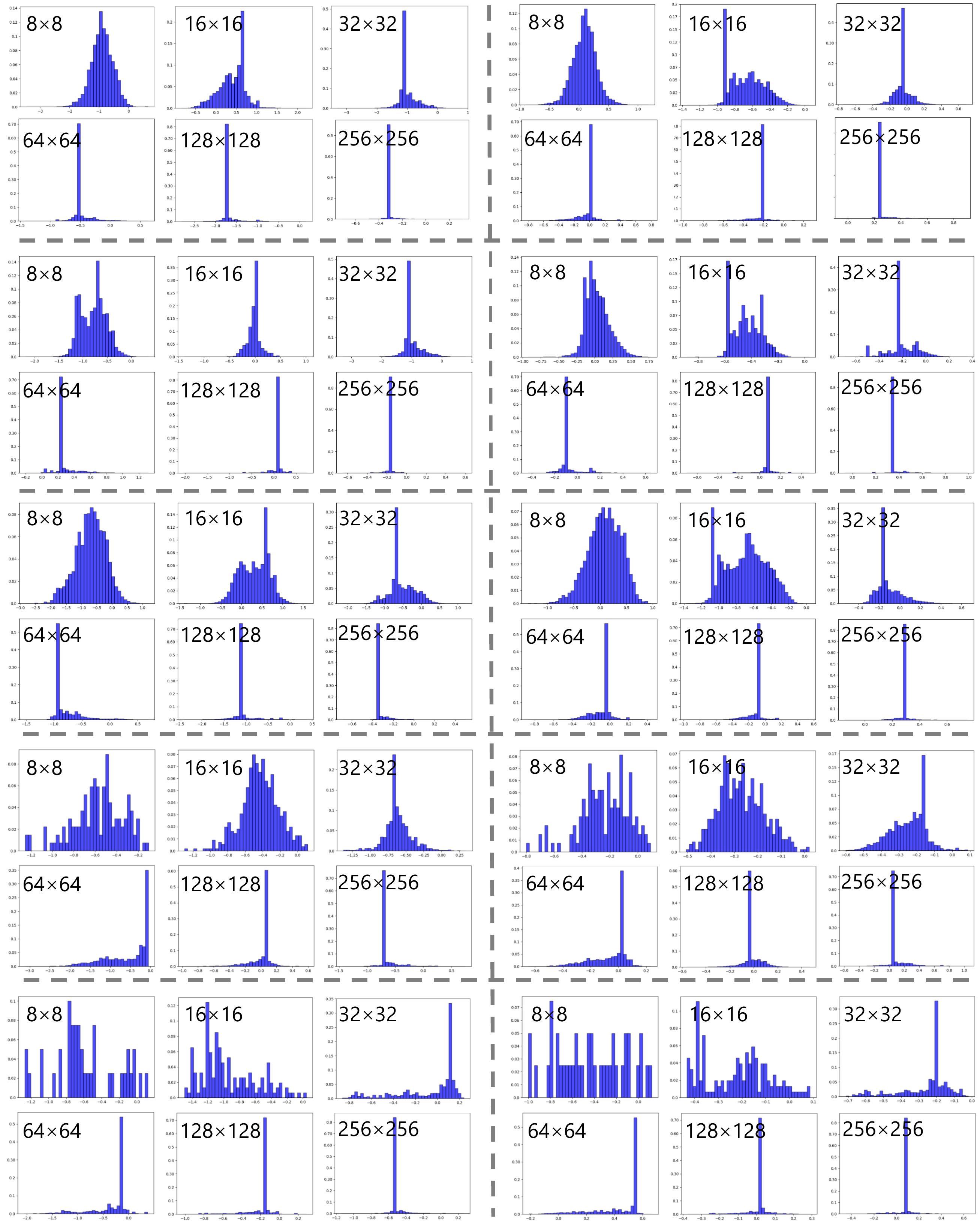}
  \caption{Statistical histograms of $\gamma$ (left) and $\beta$ (right) for the ``building'', ``sky'', ``tree'', ``human'', and ``car'' (from top to bottom) classes from the \textit{ADE20k} validation dataset on SPADE blocks with various resolutions of input masks. It can be seen that the distribution of $\gamma$ and $\beta$ is concentrated and the centralized trend becomes more obvious as the resolution of input mask goes higher.}
  \label{fig:spade-hist}
\end{figure}

As explained in \cite{park2019semantic}, the advantages of SPADE mainly come from two important properties: \emph{spatial-adaptiveness} and \emph{semantic-awareness}. The former indicates the modulation parameters $(\gamma,\beta)$ are spatially varying in a pixel-wise manner, while the latter property means that $(\gamma,\beta)$ depend on semantic classes to bring back the lost information. As the name of SPADE implies, it may indicate that the spatial-adaptiveness is more important. However, through the following analysis, we think that the semantic-awareness may be the de facto main contributor to SPADE.

In \Fref{fig:feature}, we show two examples with the masks $m$ from the \textit{ADE20k} validation dataset~\cite{zhou2017scene}, which consist of two semantic labels ``Sky'' and ``Field''. We visualize the intermediate parameters of $\gamma$ and $\beta$ with the original pre-trained SPADE generator. To show the effect clearly, we choose the input with the highest resolution in the model. It can be easily observed that $\gamma,\beta$ are almost identical within each semantic region, except for the boundary area which is especially negligible for high-resolution input masks due to the shallowness of the modulation network.
In fact, for any two regions sharing the same semantic class within one input mask or even across different input masks, their learned $\gamma, \beta$ will also be almost identical if the sizes of regions are much larger than the receptive field of the two-layer modulation network. At lower resolutions, this invariance within the semantic class will diminish. We further conduct statistical analyses of $\gamma$ and $\beta$ with the original pre-trained SPADE generator for some semantic classes on the \textit{ADE20k} validation dataset~\cite{zhou2017scene}.
In \Fref{fig:spade-hist}, we show the statistical histograms of $\gamma$ and $\beta$ for the five common classes (``building'', ``sky'', ``tree'', ``human'', and ``car'') on SPADE blocks with various resolutions of input masks. 
We can observe that the distributions of $\gamma,\beta$ within the same semantic class are concentrated and
the trend of concentration becomes more obvious as the resolution of the input mask increases.
This further proves that, compared with the spatially-adaptiveness, the semantic-awareness may be the underlying key to the superior performance of SPADE.

\begin{figure*}[tp]
  \centering
  \includegraphics[width=\linewidth]{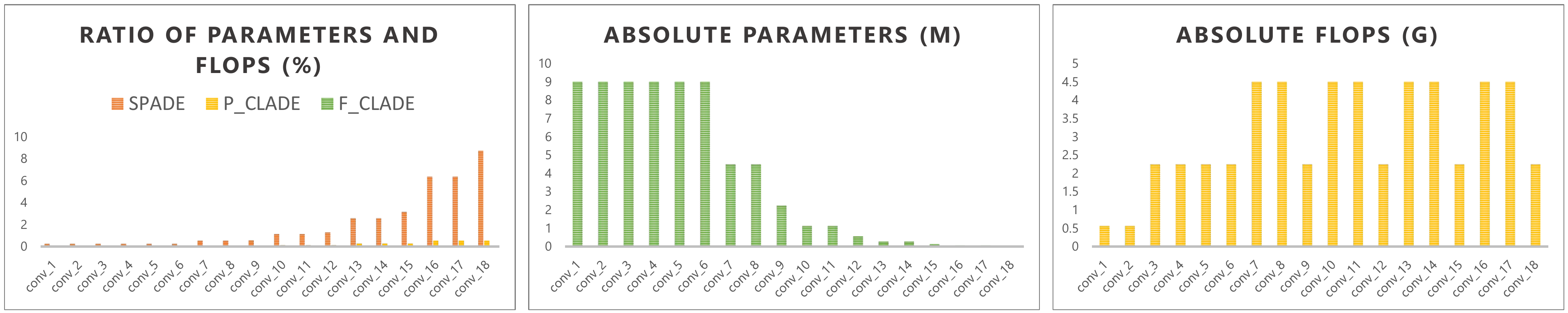}
  \caption{Left: the relative ratios of the parameter and FLOPs between each SPADE/CLADE and its following convolutional layer in the generator. The ratios of parameter and FLOPs for SPADE are the same and shown in orange, while the ratios of parameter and FLOPs for CLADE are shown in yellow and green respectively. Since the ratio of FLOPs for CLADE is very small, it is almost invisible in the figure. Middle and right: the numbers of absolute parameters and FLOPs of different convolution layers. $x-$axis indicates the layer index from deep to shallow.}
  \label{fig:ratio}
\end{figure*}

\subsection{Class-Adaptive Normalization}
Inspired by the above observation, we propose a new efficient conditional normalization layer, called CLass-Adaptive (DE)normalization (CLADE), as shown in the right of \Fref{fig:SPADE-CLADE}. Inheriting the idea of semantic information compensation from SPADE, the modulation parameters $(\gamma,\beta)$ in CLADE are also adaptive to the semantic input of $m$. However, instead of adopting the pixel-wise spatial-adaptiveness as in SPADE, CLADE is spatially-invariant and only adaptive to different semantic classes. More concretely, $(\gamma, \beta)$ in CLADE vary on the corresponding semantic classes to maintain the essential property of semantic-awareness, but they are independent of any spatial information including the position, semantic shape, or layout of $m$.

Therefore, rather than learning modulation parameters through an extra modulation network like SPADE, we directly maintain a modulation parameter bank for CLADE and optimize it as regular network parameters. Assuming the total class number in $\mathbb{L}$ to be $N_c$, the parameter bank consists of $N_c$ channel-wise modulation scale parameters $\Gamma=(\gamma^1_k,...,\gamma^{N_c}_k)$ and shift parameters $B=(\beta^1_k,...,\beta^{N_c}_k)$. During training, given an input mask $m$, we fill each semantic region of class $l$ with its corresponding modulation parameter $\gamma_k^l, \beta_k^l$ to generate dense modulation parameter tensors $\overrightarrow{\gamma}$ and $\overrightarrow{\beta}$ respectively. We call this process \emph{Guided Sampling} in \Fref{fig:SPADE-CLADE}.

In fact, CLADE can also be regarded as a generalized formulation of some existing normalization layers. If $\gamma_k^{l_1}\equiv\gamma_k^{l_2}$ and  $\beta_k^{l_1}\equiv\beta_k^{l_2}$ for any $l_1,l_2\in\mathbb{L}$, CLADE becomes BN~\cite{ioffe2015batch}. And if we make the modulation tensors $\overrightarrow{\gamma}$ and $\overrightarrow{\beta}$ both spatially uniform, and replace the mean and std statistics of BN with the corresponding ones from IN, we implement Conditional IN. 

By default, CLADE uses the additional input of instance maps if provided by the datasets (Cityscapes and COCO-Stuff) to better distinguish the different instances of the same categories. Similar to pix2pixHD and SPADE, we feed the edge map $E$ calculated from the instance map (‘edge’ and ‘non-edge’ are represented as ‘1’ and ‘0’) into the network. However, the special architecture of CLADE does not allow us to stack the edge map with the semantic layout directly. Thus, we embed the edge information in the modulated features. To match the activation values in the feature, we first modulate the edge map as follows:
\begin{equation}
    \hat{E} = \gamma_{c} * E + \beta_{c},
\label{eq:edge}
\end{equation}
where $\hat{E}$ is the modulated edge map. $\gamma_c$ and $\beta_c$ are two constant float point numbers that can be learned as regular parameters. Then, we combine the modulated $\hat{E}$ with the feature maps $x^{out}$ modulated by the CLADE layer along the channel dimension, and feed them into the following layers. Since only two constant numbers are involved and \Eref{eq:edge} can also be implemented by pixel-wise value assignment operations, the extra parameter and computation overhead is extremely low and negligible.

\subsection{Computation and Parameter Complexity Analysis}

\subsubsection{Analysis of SPADE}
In the original SPADE generator backbone~\cite{park2019semantic}, a SPADE block is placed before almost every convolution to replace the conventional normalization layer. For convenience, we denote the input and output channel numbers of the following convolutional layer as $C_{in}, C_{out}$ and its kernel size as $k_c$. For its modulation network, we simply assume a same kernel size $k_m$ and intermediate channel number $C_m$ are used for all convolutional layers. Therefore, the parameter numbers for the convolutional layer $P_{conv}$ and the SPADE block $P_{spade}$ are calculated as:
\begin{equation}
    P_{conv} = k_c^2*C_{in}*C_{out},
\end{equation}
\begin{equation}
    P_{spade} = k_m^2*(N_c*C_m + 2*C_m*C_{in}).
\end{equation}

With the default implementation settings of SPADE, we have $k_c=k_m=3$, so the parameter ratio between both networks is:
\begin{equation}
P_{spade}/P_{conv}=\frac{N_c*C_m + 2*C_m*C_{in}}{C_{in}*C_{out}}.
\end{equation}
This to say, the extra parameter introduced by SPADE becomes a significant overhead, especially when $N_c,C_m$ are relatively large ($C_m=128$ by default in SPADE). Take the \textit{ADE20k} dataset~\cite{zhou2017scene} as an example, which contains 151 classes ($N_c=151$). On image resolution of $256\times 256$, the SPADE generator consists of 7 SPADE residual blocks. We show the parameter ratio $P_{spade}/P_{conv}$ of each convolutional layer in \Fref{fig:ratio}. It can be seen that SPADE indeed brings considerable parameter overhead to all the convolutional layers. This becomes even more serious when the network goes deeper since $C_{out}$ is designed to be smaller for higher feature resolution. The ratios for some layers even
exceed $600\%$. Taking all the convolutional layers in SPADE generators into consideration, the average ratio is about $39.21\%$.

In addition to the parameter numbers, we also analyze the computation complexity. Here, we use the popular floating-point operation per second (FLOPs) as the metric. Since the convolutional layers within the modulation network dominate the computation cost of SPADE, the FLOPs of both the convolutional layer $F_{conv}$ and the SPADE block $F_{spade}$ can be simply calculated as:
\begin{equation}
    F_{conv} = k_c^2*C_{in}*C_{out}*H*W,
\end{equation}
\begin{equation}
    F_{spade} = k_m^2*(N_c*C_m + 2*C_m*C_{in})*H*W,
\end{equation}
where $H,W$ are the width and height of the output feature respectively. Therefore, the FLOPs ratio $F_{spade}/P_{conv}$ is identical to the parameter ratio shown in \Fref{fig:ratio}. However, different from the parameter number, with the increasing feature resolutions, the absolute FLOPs are relatively larger in shallower layers, which makes the computation overhead even worse. Taking the same \textit{ADE20k} dataset as an example, the average extra FLOPs ratio introduced by SPADE is about $234.73\%$, which means the computation cost of SPADE is even heavier than the convolutional layers. More importantly, it is now popular to adopt very large synthesis networks to ensure good performance, which is already consuming a surprisingly large amount of parameter space and computation resources, and SPADE will further aggravate this situation, which might be unaffordable in many cases.

\subsubsection{Analysis of CLADE} 
\label{sec:anl_clade}
Compared to SPADE, our CLADE does not require any extra modulation network to regress the modulation parameters. Specifically, the corresponding numbers of its parameters and FLOPs are:
\begin{equation}
    P_{clade} = 2*N_c*C_{in},
\end{equation}
\begin{equation}
    F_{clade} =2*C_{in}*H*W.
\end{equation}
We take the value assignment operation as one float-point operation. Similar to SPADE, if every convolutional layer is followed by one CLADE layer, the relative ratios of parameter and FLOPs are:
\begin{equation}\label{eq:rp_clade}
    P_{clade}/P_{conv}=\frac{2*N_c}{k_c^2*C_{out}}, 
\end{equation}
\begin{equation}\label{eq:rf_clade}
    F_{clade}/F_{conv}=\frac{2}{k_c^2*C_{out}}.
\end{equation}
In most existing synthesis networks, the above ratios are extremely small. For example, with the same backbone as the above SPADE generator for the \textit{ADE20k} dataset, the ratios of parameter and FLOPs for each CLADE layer are much less than those of SPADE (shown on the left of \Fref{fig:ratio}). Finally, the average ratios for parameter and FLOPs are only $4.57\%$ and $0.07\%$, respectively. Therefore, compared to SPADE, the parameter and computation overhead of CLADE are negligible, which is friendly to practical scenarios regarding both training and inference. Despite its simplicity and efficiency, we demonstrate that it can still achieve comparable performance as SPADE with extensive experiments in \Sref{sec:exp}.

\begin{figure}[t]
  \centering
  \includegraphics[width=0.85\linewidth]{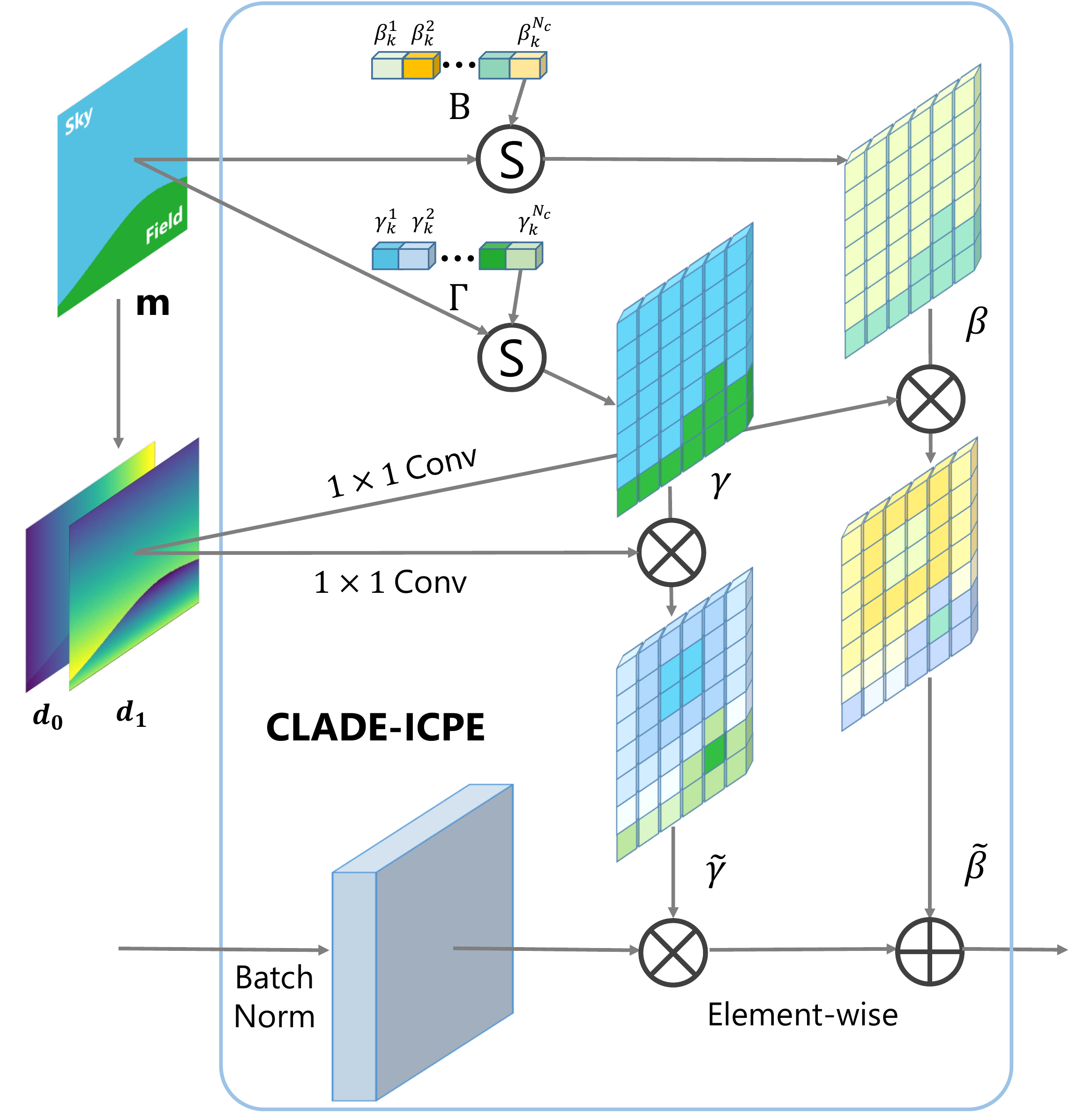}
  \caption{The illustration of class-adaptive normalization layer (CLADE) with intra-class positional encoding (ICPE). The positional encoding map is calculated from the semantic segmentation map. $d_0$ and $d_1$ represent the positional encoding along the $x,y$ dimension.}
  \label{fig:dist}
\end{figure}
\subsection{Spatially-Adaptive CLADE}
\label{sec:ICPE}
As mentioned before, the modulation parameters of SPADE are almost spatially invariant within the same semantic region, especially for high-resolution input masks. In other words, the spatial-adaptiveness is not fully utilized in SPADE. This is mainly due to the limited receptive fields of modulation layers in a shallow network. Theoretically, if we increase the depth of the network, better spatial-adaptiveness could be achieved with the accumulation of receptive fields, but along with prohibitively high computational cost. Based on this observation, we propose a variant of CLADE, CLADE-ICPE, to further improve intra-class spatial adaptiveness by leveraging a positional encoding map as the extra input.

The positional encoding map is defined as the relative distance from each pixel to its corresponding object center, which can be calculated using the input semantic mask $m$. Specifically, for each pixel $(i,j)$ in the positional encoding map $d\in \mathbb{R}^{2\times H\times W}$, we first find its belonging semantic object ($o_l$) by detecting the largest connected component of the corresponding semantic category $l$ and obtain the object center ($cx_{o_l},cy_{o_l}$). Then the distance map along the $x,y$ dimension is defined as:
\begin{equation}
    d_{i,j,0}^{'} = (i - cx_{o_l}), \quad d_{i,j,1}^{'} = (j - cy_{o_l}).
\end{equation}
We further define the maximum offset of each object $o_l$ as:
\begin{equation}
    mo_{k} = \arg\max_{i,j} d_{i,j,k}^{'}, \quad k=\{0,1\}, \quad \text{s.t}. \; (i,j)\in o_l.
\end{equation}
Finally, we get the normalized distance map $d$ by normalizing with the maximum offset:
\begin{equation}
    d_{i,j,0} = d_{i,j,0}^{'}/mo_0, \quad d_{i,j,1} = d_{i,j,1}^{'}/mo_1, \quad \text{s.t}. \; (i,j)\in o_l.
\end{equation}
As shown in \Fref{fig:dist}, in order to utilize the positional encoding map $d$, we follow the modulation idea and use a $1\times 1$ convolution layer to map the positional encoding to the modulation parameters ($\gamma / \beta$):
\begin{equation}
    \widetilde{\gamma} = \gamma \otimes (1+\mathcal{C}_{\gamma}(d)), \quad \widetilde{\beta} = \beta \otimes (1+\mathcal{C}_{\beta}(d)),
\end{equation}
where $\mathcal{C}_{\gamma}$ and $\mathcal{C}_{\beta}$ are convolution operations with one-channel outputs. And $\otimes$ is the element-wise multiplication. Since the input and output channel numbers of $\mathcal{C}_{\gamma}$ and $\mathcal{C}_{\beta}$ are $2$ and $1$, respectively, the extra parameter and computation overhead is almost negligible. Specifically, the corresponding relative ratios of parameters an FLOPs defined in \Sref{sec:anl_clade} are:
\begin{equation}
    P_{clade-icpe}/P_{conv}=\frac{2*N_c+4/C_{in}}{k_c^2*C_{out}}, 
\end{equation}
\begin{equation}
    F_{clade-icpe}/F_{conv}=\frac{4+4/C_{in}}{k_c^2*C_{out}}.
\end{equation}
Compared with \Eref{eq:rp_clade}, the ratio of parameters is almost the same, while the ratio of FLOPs is almost twice that of CLADE. However, the absolute ratio is still relatively low, especially compared to SPADE (0.14\% vs. 234.73\%).

\begin{figure*}[tp]
  \centering
  \includegraphics[width=\linewidth]{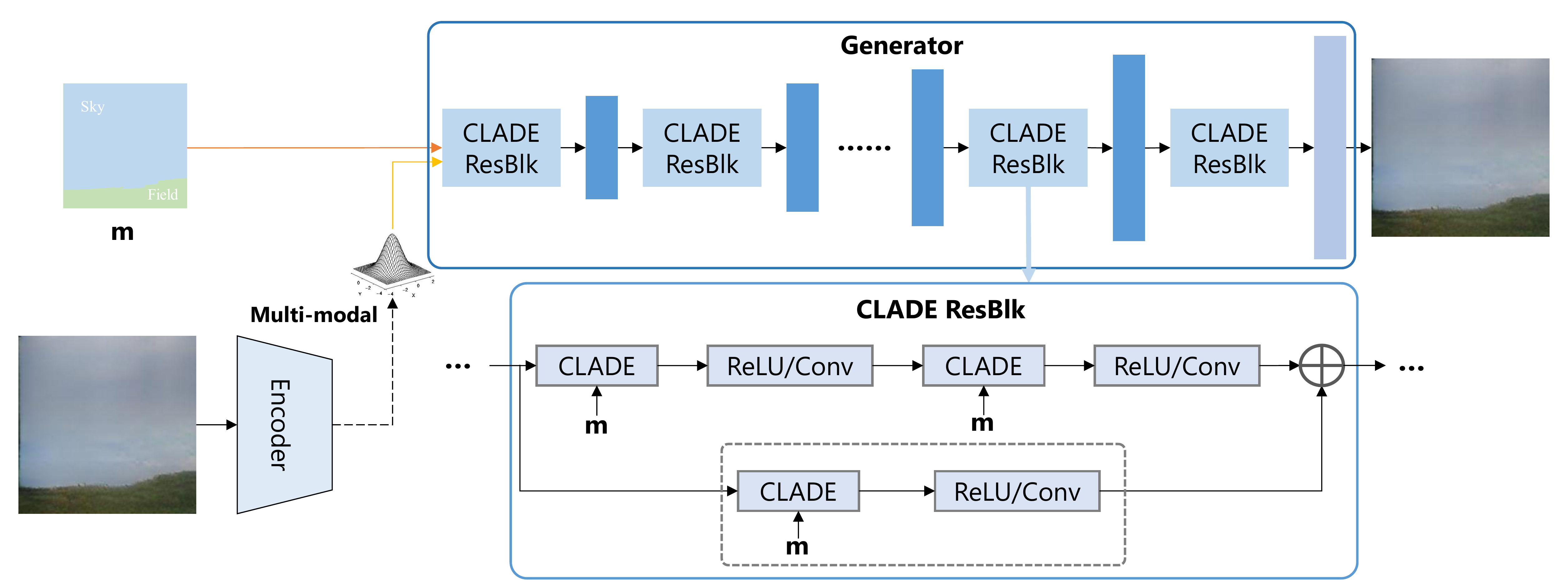}
  \caption{Architecture of our generator. By default, we feed the downsampled semantic mask to the generator. When processing multi-modal image generation, the input of generator is replaced by a random noise. For style-guided synthesis, a style encoder is used to guide the specified distribution.}
  \label{fig:generator}
\end{figure*}

\subsection{CLADE Generator}
Similar to SPADE, our proposed CLADE can be integrated into different generator backbones. In this paper, the CLADE generator follows the similar network architecture of the SPADE generator \cite{park2019semantic} by default, but all the SPADE blocks are replaced by CLADE. As shown in \Fref{fig:generator}, it adopts several residual blocks with upsampling layers and progressively increases the output resolution. The residual block consists of CLADE layers, ReLU layers and convolution layers, and the skip connection is also replaced by these layers when the number of channels before and after the residual block are different.
For multi-modal synthesis, we follow the strategy as~\cite{park2019semantic} and attach an extra encoder that encodes the image into a random vector. Specifically, this encoder consists of a series of convolutional layers with stride 2, instance normalization layers and LReLU activation layers and outputs the mean and variance vector of the distribution of the specified image. Then a random vector sampled from this distribution is fed into the CLADE generator as the style guidance to enable global diversity of the generated results.

\section{Experiments}
\label{sec:exp}

\subsection{Datasets} Main experiments are conducted on four popular datasets: \textit{ADE20k}, \textit{ADE20k-outdoor}, \textit{COCO-Stuff}, and \textit{Cityscapes}. The \textit{ADE20k} dataset~\cite{zhou2017scene} consists of 25,210 images (20,210 for training, 2,000 for validation and 3,000 for testing), covering a total of 150 object and stuff categories.  \textit{ADE20k-outdoor} is a subset of \textit{ADE20k} that only contains outdoor scenes. Similar to previous work~\cite{park2019semantic,qi2018semi}, we directly select the images containing categories such as sky, trees, and sea without manual inspection. There are 9,649 training images and 943 validation images. The \textit{COCO-Stuff} dataset~\cite{caesar2018coco} augments \textit{COCO} by adding dense pixel-wise stuff annotations. It has 118,000 training images and 5,000 validation images with 182 semantic categories. The \textit{Cityscapes} dataset~\cite{cordts2016cityscapes} is a widely used dataset for semantic image synthesis. It contains 2,975 high-resolution training images and 500 validation images of 35 semantic categories. 

We use two additional datasets to evaluate the generalization ability when applying our CLADE to some recent semantic synthesis methods that depend on SPADE. The \textit{CelebAMask-HQ}~\cite{lee2020maskgan,karras2017progressive,liu2015deep} contains 30,000 segmentation masks with 19 different classes from CelebAHQ face imgae dataset. They are split into 28,000 training images and 2,000 validation images. The \textit{DeepFashion}~\cite{liu2016deepfashion} contains  52,712 person images with fashion clothes. We use the processed dataset provided by SMIS~\cite{zhu2020semantically} which consists of 30,000 training images and 2,247 validation images. 

\begin{table*}[t]
    \centering
    \caption{Performance and complexity comparison with other semantic image synthesis methods. All the metrics are tested by ourselves on the PyTorch and Titan XP GPU.}
    \small
    \setlength{\tabcolsep}{1.8mm}
    \begin{tabular}{c|c|c|c|c|c|c|c|c}
    \hline
        Dataset & Method & mIoU $\uparrow$ & accu $\uparrow$ & FID $\downarrow$ & Params (M) $\downarrow$ & FLOPs (G) $\downarrow$ & Runtime (s) $\downarrow$ & Memory (GB) $\downarrow$\\
        \hline
         
         \multirow{3}{*}{ADE20k} & CC-FPSE & \textbf{38.93} & \textbf{79.75} & 29.74 & 140.7 & 438.2 & 0.093 & -\\
         & DAGAN & 38.07 & 79.52 & 28.93 & 96.6 & 186.1 & 0.057 & 4.36\\
         & pix2pixHD & 27.27 & 72.61 & 45.87 & 182.9 & 99.3 & 0.041 & 7.47\\
         & SPADE & 36.28 & 78.13 & 29.79 & 96.5 & 181.3 & 0.042 & 4.36\\
         & CLADE & 35.43 & 77.36 & 30.48 & \textbf{71.4} & \textbf{42.2} & \textbf{0.024} & \textbf{3.68}\\
         & CLADE-ICPE & 35.06 & 77.09 & \textbf{28.69} & \textbf{71.4} & \textbf{42.2} & 0.027 & 4.00\\
        \hline
        \multirow{3}{*}{ADE20k-outdoor} & pix2pixHD & 14.89 & 76.70 & 67.13 & 182.9 & 99.3 & 0.041 & 7.47\\
         & SPADE & \textbf{19.30} & 80.44 & 45.92 & 96.5 & 181.3 & 0.042 & 4.36\\
         & CLADE & 18.71 & \textbf{80.77} & 46.37 & \textbf{71.4} & \textbf{42.2} & \textbf{0.024} & \textbf{3.68}\\
         & CLADE-ICPE & 18.89 & 80.04 & \textbf{45.59} & \textbf{71.4} & \textbf{42.2} & 0.027 & 4.00\\
        \hline
         \multirow{3}{*}{COCO-Stuff} & CC-FPSE & \textbf{40.40} & \textbf{69.64} & \textbf{25.39} & 141.9 & 456.1 & 0.094 & -\\
         & pix2pixHD & 21.07 & 54.80 & 58.52 & 183.0 & 106.1 & 0.046 & 8.32\\
         & SPADE & 36.74 & 67.81 & 27.69 & 97.5 & 191.3 & 0.046 & 4.62\\
         & CLADE & 36.77 & 68.08 & 29.16 & \textbf{72.5} & \textbf{42.4} & \textbf{0.027} & \textbf{3.85}\\
         & CLADE-ICPE & 36.39 & 67.57 & 27.76 & \textbf{72.5} & \textbf{42.4} & 0.030 & 4.06\\
        \hline
        \multirow{3}{*}{Cityscapes}
         & CC-FPSE & \textbf{65.48} & \textbf{93.84} & 43.69 & 128.1 & 739.2 & 0.157 & -\\
         & DAGAN & 63.64 & 93.47 & 55.63 & 93.1 & 291.2 & 0.132 & 5.46\\
         & pix2pixHD & 60.50 & 93.06 & 66.04 & 182.5 & 151.3 & 0.038 & 4.88\\
         & SPADE & 61.95 & 93.39 & 51.98 & 93.0 & 281.5 & 0.065 & 5.46\\
         & CLADE & 60.44 & 93.42 & 50.62 & \textbf{67.9} & \textbf{75.5} & \textbf{0.035} & \textbf{4.37}\\
         & CLADE-ICPE & 60.40 & 93.26 & \textbf{42.39} & \textbf{67.9} & \textbf{75.5} & 0.039 & 4.85\\
        \hline
    \end{tabular}
    \label{tab:performance}
\end{table*}{}

\begin{table}[t]
    \centering
    \caption{Runtime comparison between SPADE and CLADE on a single-threaded CPU (Intel(R) Xeon(R) Gold 6148 CPU @ 2.40GHz).}
    \begin{tabular}{c|c|c|c|c}
    \hline
        \multirow{2}{*}{Method} & \multirow{2}{*}{ADE20k} & ADE20k & \multirow{2}{*}{COCO-Stuff} & \multirow{2}{*}{Cityscapes} \\
         & & -outdoor & & \\
        \hline
        SPADE & 4.104s & 4.104s & 4.226s & 7.031s \\
        \hline
        CLADE & 1.710s & 1.710s & 1.875s & 2.936s \\
        \hline
    \end{tabular}
    \label{tab:cpu_time}
\end{table}

\begin{table*}[t]
    \centering
    \caption{Detailed comparison with SPADE and CLADE on the ADE20k (Col 2-4), Cityscapes (Col 5-7) and COCO-Stuff (Col 8-10) datasets. Backbone represents the generator without normalization layers, SPADE and CLADE represent the different normalization layers.}
    \begin{tabular}{c||c|c|c||c|c|c||c|c|c}
        \hline
        Model & Backbone & SPADE & CLADE &  Backbone & SPADE & CLADE &  Backbone & SPADE & CLADE\\
        \hline
        Params (M) $\downarrow$ &  68.1 & 28.4 & \textbf{3.3} & 67.1 & 25.9 & \textbf{0.8} & 68.4 & 29.1 & \textbf{4.1} \\
        Runtime (s) $\downarrow$ &  0.015 & 0.027 & \textbf{0.009} & 0.022 & 0.043 & \textbf{0.013} & 0.017 & 0.029 & \textbf{0.010} \\
        \hline
    \end{tabular}
    \label{tab:cmp_time}
\end{table*}

\begin{table*}[t]
    \centering
    \caption{Performance comparison with a lightweight model of SPADE on four datasets. The compared methods have the similar  FLOPs.}
    \small
    \begin{tabular}{c|cccc|cccc}
    \hline
        \multirow{2}{*}{Dataset} & \multicolumn{4}{c|}{SPADE-light} & \multicolumn{4}{c}{CLADE}\\
        \cline{2-9}
         & mIoU $\uparrow$ & accu $\uparrow$ & FID $\downarrow$ & FLOPs (G) $\downarrow$ & mIoU $\uparrow$ & accu $\uparrow$ & FID $\downarrow$ & FLOPs (G) $\downarrow$\\
        \hline
        ADE20k & 26.29 & 71.76 & 40.45 & 58.0 & \textbf{35.43} & \textbf{77.36} & \textbf{30.48} & \textbf{42.2}\\
        \hline
        ADE20k-outdoor & 15.54 & 77.69 & 58.55 & 58.0 & \textbf{18.71} & \textbf{80.77} & \textbf{46.37} & \textbf{42.2}\\
        \hline
        COCO-Stuff & 27.01 & 60.64 & 44.19 & 68.0 & \textbf{36.77} & \textbf{68.08} & \textbf{29.16} & \textbf{42.4}\\
        \hline
        Cityscapes & 59.70 & 93.13 & 52.07 & 132.9 & \textbf{60.44} & \textbf{93.42} & \textbf{50.62} & \textbf{75.5}\\
        \hline
    \end{tabular}
    \label{tab:spade-light}
\end{table*}{}

\subsection{Implementation Details}
We follow the same training setting as SPADE~\cite{park2019semantic}. In details, the generator is trained with the same multi-scale discriminator and the loss function is as follows:
\begin{equation}
    \mathcal{L} = \mathcal{L}_{GAN}+\lambda_{1} \mathcal{L}_{FM}+ \lambda_{2} \mathcal{L}_{P},
\end{equation}
where $\mathcal{L}_{GAN}$ is the hinge version of GAN loss, and $\mathcal{L}_{FM}$ is the feature matching loss between the real and synthesized images. The feature is extracted by the multi-scale discriminator. $\mathcal{L}_{P}$ is the perceptual loss~\cite{johnson2016perceptual} with the feature extractor of VGG network~\cite{simonyan2014very}. For multi-modal synthesis, we add KL-divergence loss term ($\lambda_{3} \mathcal{L}_{KL}$) to minimize the gap between the encoded distribution and Gaussian distribution. By default, we set $\lambda_{1}=10, \lambda_{2}=10, \lambda_{3}=0.05$, and the Adam optimizer~\cite{kingma2015adam} ($\beta_1=0,\beta_2=0.9$) is used with the total epoch number of 200. The learning rates for the generator and discriminator are set to 0.0001 and 0.0004, respectively. We evaluate the model every 10 epochs and select the model with the best performance.
To demonstrate the effectiveness of our method, we not only compare our CLADE with the baseline of SPADE~\cite{park2019semantic} but also include the comparison with the popular semantic image synthesis method pix2pixHD~\cite{wang2018high} and two recent methods: CC-FPSE~\cite{liu2019learning} and DAGAN~\cite{tang2020dual}.
For pix2pixHD, we use the codes and settings provided by the authors to train all the models. For SPADE, CC-FPSE and DAGAN, we directly use the pre-trained models provided by the authors to generate the results for evaluation.
The resolution of images ($H\times W$) is set to $256\times256$ except for \textit{Cityscapes}, which is set to $256\times512$.

\subsection{Evaluation Metrics} We leverage the protocol from previous works~\cite{chen2017photographic,wang2018high} for evaluation, which is also used in SPADE~\cite{park2019semantic}. Specifically, we run semantic segmentation algorithms on the synthesized images and evaluate the quality of the predicted semantic masks. 
To measure the segmentation accuracy, two popular metrics, mean Intersection-over-Union (mIoU) and pixel accuracy (accu) metrics are used. For different datasets, we select corresponding state-of-the-art segmentation models: UperNet101~\cite{xiao2018unified,github-upernet} for \textit{ADE20k} and \textit{ADE20k-outdoor}, DeepLabv2~\cite{chen2017deeplab,github-deeplab} for \textit{COCO-Stuff}, DRN~\cite{yu2017dilated,github-drn} for \textit{Cityscapes} and UNet~\cite{ronneberger2015u,github-celebamask} for \textit{CelebAMask-HQ}. As for \textit{DeepFashion}, we also use UNet but train the model by ourselves.
We also leverage the commonly used Fr\'echet Inception Distance (FID)~\cite{heusel2017gans} to measure the distribution distance between synthesized images and real images. Specifically, we calculate FID between generated validation images and real training images, not generated validation images and real validation images. This is because the number of training images is more than of validation images, which can better reflect the distribution characteristics of real images. The same protocol is also adopted in the recent work\cite{choi2020stargan}.

\subsection{Quantitative Results} 
\label{sec:quantitative}
As shown in \Tref{tab:performance}, our method can achieve comparable performance with SPADE while significantly reducing the parameter number and computational complexity of the original SPADE generator on all the datasets. For example, on the \textit{COCO-Stuff} dataset, the proposed CLADE achieves a mIoU score of 36.77 and a pixel accuracy score of 68.08, which is even slightly better than SPADE. When compared to pix2pixHD, CLADE outperforms it by more than $15$ and $13$ points in terms of mIoU and pixel accuracy respectively. As for the FID score, our CLADE is also close to SPADE and much better than pix2pixHD. On the \textit{Cityscapes} dataset, our CLADE performs better than SPADE in terms of FID, but the parameter number in our CLADE generator is only about $74\%$ of that in the original SPADE generator and $39\%$ of that in pix2pixHD. As for the computation complexity in terms of FLOPs, CLADE generator is about $4\times$ fewer than that in the SPADE generator and $2\times$ fewer than that in the pix2pixHD. 
We also compare with the state-of-the-art method CC-FPSE. Although it achieves better performance than both our CLADE and SPADE, it causes much more computation cost (around $2\times$, $10\times$ and $4\times$ heavier than our CLADE in terms of parameter number, FLOPs and runtime, respectively). Besides, we compare the memory usage during training in \Tref{tab:performance} when the batch size is set to 1. By using the class-adaptive normalization, CLADE requires much less memory than the other methods. Moreover, this advantage will be more pronounced when a larger batch size is used. In other words, we can train the model with larger batch sizes on the same GPU devices.

In terms of the runtime, since the GPU computation capacity is often overqualified for single image processing, the real runtime speedup is less significant than FLOPs, but we still observe about $2\times$ speedup when compared to SPADE. We further compare the runtime on a single-threaded CPU in \Tref{tab:cpu_time}, which shows a more significant speedup. This indicates that our CLADE has more advantages when deployed on low-end devices.

Taking one step forward, we further analyze the extra parameter and computation cost introduced by SPADE and CLADE in \Tref{tab:cmp_time}. In details, we calculate the parameter and computation cost brought by the backbone network (operations except normalization) and the SPADE (or CLADE) layers respectively. It can be seen that in \Tref{tab:cmp_time}, the advantages of CLADE layers in terms of parameters and runtime are much more obvious when ignoring the backbone part. 

When introducing additional spatial information, CLADE-ICPE has made a significant improvement in terms of FID on all the datasets. Even compared to SPADE, CLADE-ICPE shows a considerable advantage, especially on \textit{Cityscapes} dataset. But as for the model complexity, the additional parameters and FLOPs are negligible, and the overhead increase in the average running time and training memory is also small.

To further demonstrate the efficiency and effectiveness of CLADE, we also train a lightweight variant of SPADE (denoted as SPADE-light in \Tref{tab:spade-light}) by reducing the number of channels in its convolution layers to ensure it has similar FLOPs as CLADE. Obviously, SPADE-light performs much worse than CLADE on all datasets.

\begin{table}[tp]
    \centering
    \caption{User study results. The numbers indicate the percentage of users who favor the results of the proposed CLADE over the competing method.}
    \begin{tabular}{c|c|c|c|c}
    \hline
        \multirow{2}{*}{Method} & \multirow{2}{*}{ADE20k} & ADE20k & \multirow{2}{*}{COCO-Stuff} & \multirow{2}{*}{Cityscapes} \\
         & & -outdoor & & \\
        \hline
        CLADE vs.  & \multirow{2}{*}{48.375} & \multirow{2}{*}{57.000} & \multirow{2}{*}{55.000} & \multirow{2}{*}{53.375} \\
        SPADE & & & &  \\
        \hline
        CLADE vs.  & \multirow{2}{*}{68.375} & \multirow{2}{*}{73.375} & \multirow{2}{*}{95.000} & \multirow{2}{*}{57.500} \\
        pix2pixHD & & & &  \\
        \hline
        CLADE vs.  & \multirow{2}{*}{30.375} & \multirow{2}{*}{42.500} & \multirow{2}{*}{48.750} & \multirow{2}{*}{25.000} \\
        CLADE-ICPE & & & &  \\
        \hline
    \end{tabular}
    \label{tab:userstudy}
\end{table}

\subsection{User Study} Since judging the visual quality of one image is usually subjective, we further conduct a user study to compare the results generated by different methods. Specifically, we give the users two synthesis images generated from the same semantic mask by two different methods (our method and the baseline method) and ask them ``which is more realistic''. To ensure a more detailed comparison, there is no time limit set for the users. And for each pairwise comparison, we randomly choose 40 results for each method and involve 20 users. In \Tref{tab:userstudy}, we report the evaluation results on four different datasets. According to the results, we find that users have no obvious preference between our CLADE and SPADE, which once again proves the comparable performance to SPADE. But compared to the results of pix2pixHD, users clearly prefer our results on all the datasets, especially including the challenging \textit{COCO-Stuff} dataset. When comparing the results of CLADE and CLADE-ICPE, users prefer the latter, especially for the results on the Cityscapes dataset. As for the results on the COCO-Stuff dataset, it seems that users can hardly decide which one is better. But in general, it shows that CLADE-ICPE can generate better visual results than CLADE, which is consistent with FIDs in \Tref{tab:performance}.

\subsection{Qualitative Results}
\begin{figure}[tp]
  \centering
  \includegraphics[width=\linewidth]{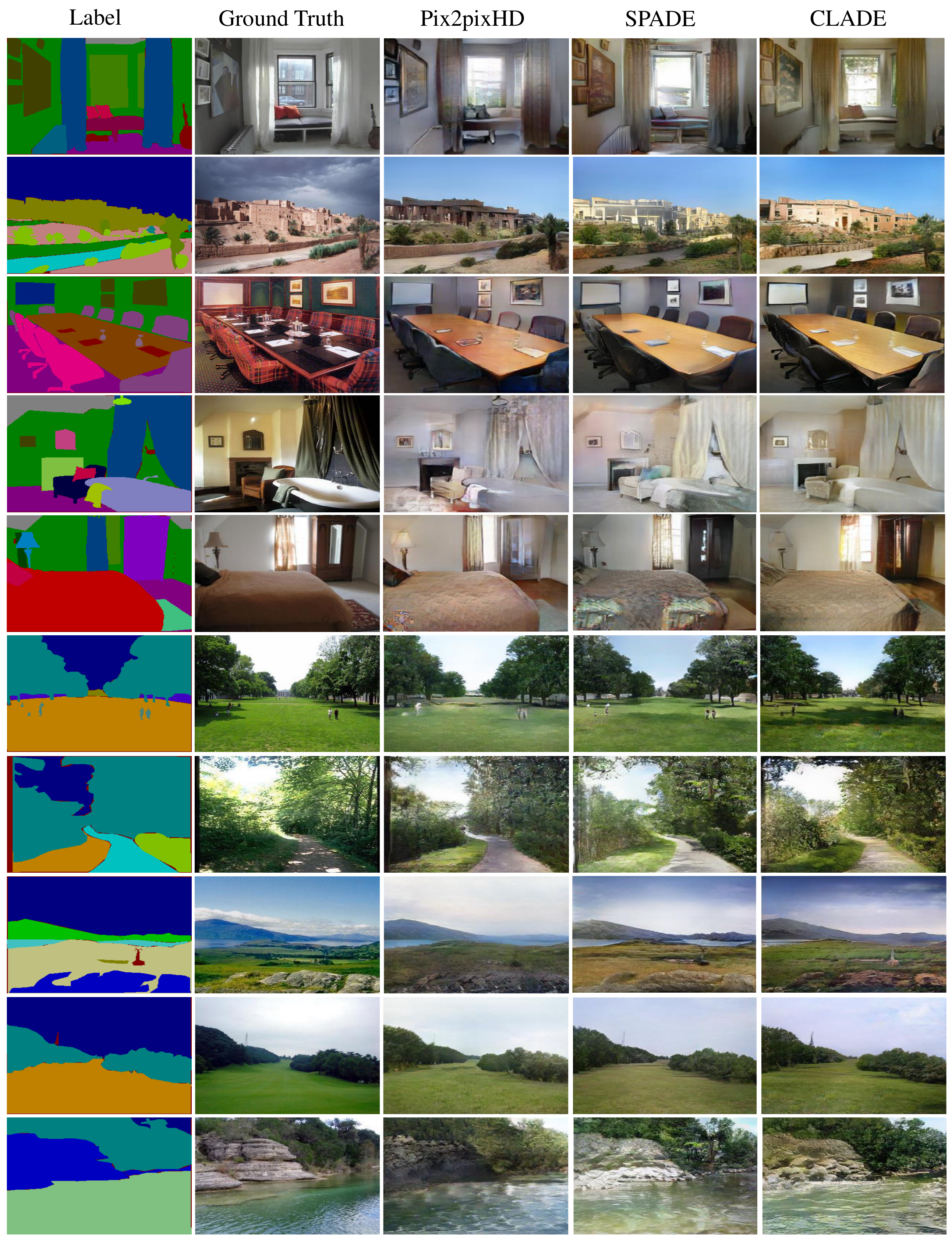}
  \caption{Visual comparison results on the \textit{ADE20k} (top five rows) and \textit{ADE20k-outdoor} (bottom five rows) dataset. It shows that images generated by our method are very comparable or even slightly better than SPADE. Compared to Pix2pixHD, SPADE and CLADE are overall more realistic.}
  \label{fig:results_ade}
\end{figure}

\begin{figure}[tp]
  \centering
  \includegraphics[width=\linewidth]{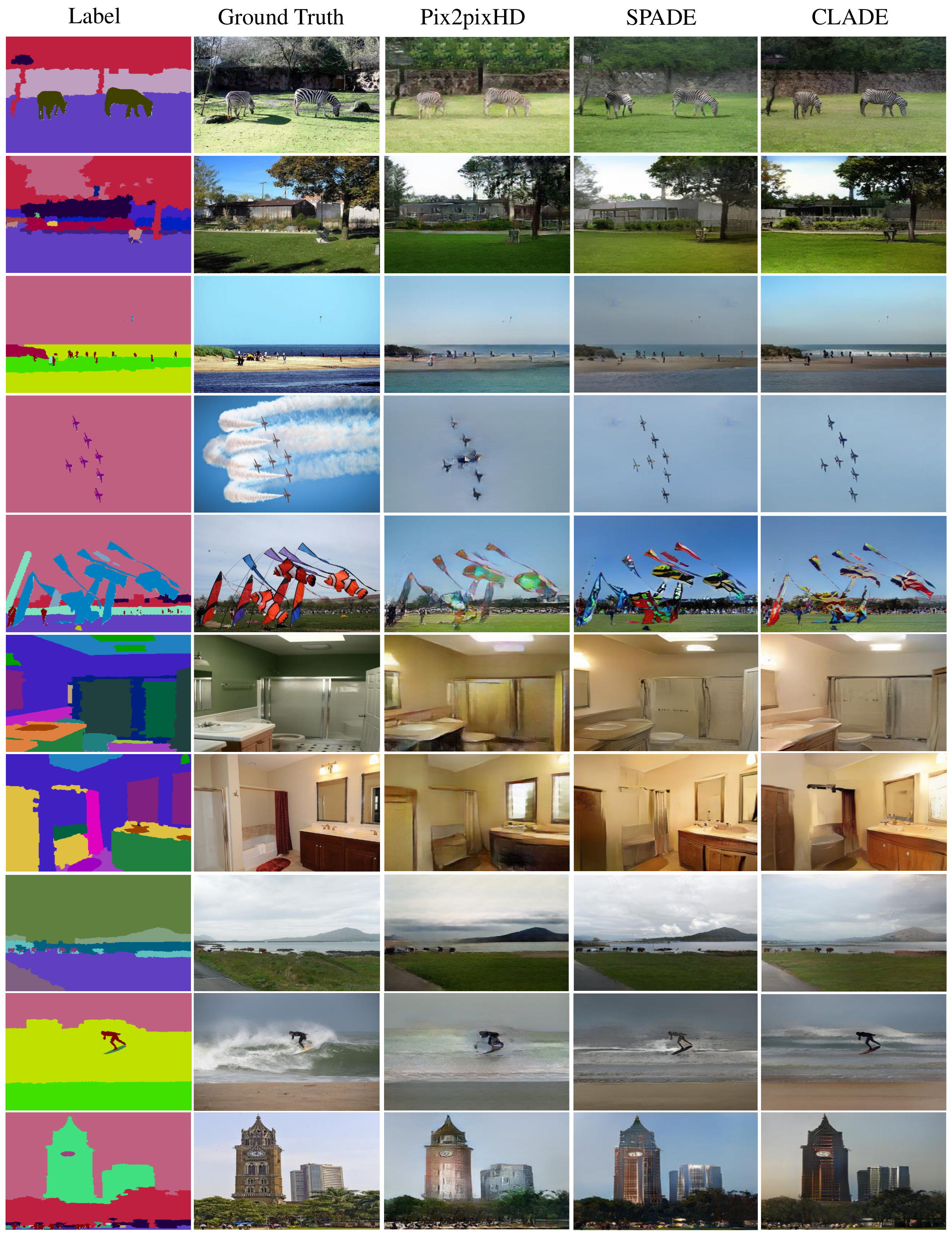}
  \caption{Visual comparison results on the challenging \textit{COCO-Stuff} dataset. Though very diverse categories and small structures exist in this dataset, our method can work very well and generate very high-fidelity results.}
  \label{fig:results_coco}
\end{figure}

\begin{figure}[tp]
  \centering
  \includegraphics[width=\linewidth]{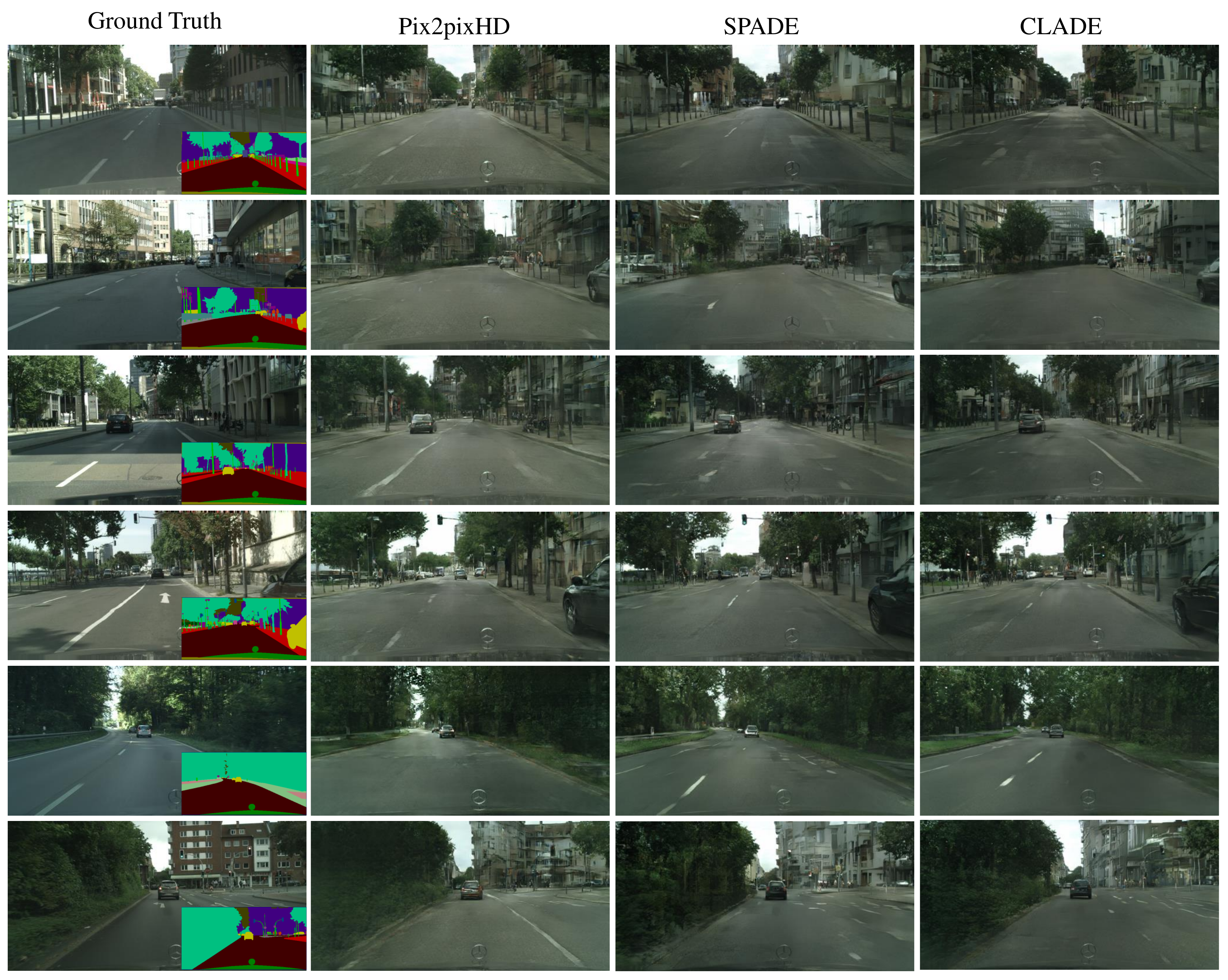}
  \caption{High-resolution synthesis ($256\times 512$) results on the \textit{Cityscapes} dataset. Our method produces realistic images with faithful spatial alignment and semantic meaning.}
  \label{fig:results_city}
\end{figure}

Besides the above quantitative comparison, we further provide some qualitative comparison results on the four different datasets. In detail, \Fref{fig:results_ade} shows some visual results on some indoor cases on the \textit{ADE20k} dataset and outdoor cases on the \textit{ADE20k-outdoor} dataset. Despite the simplicity of our method, it can generate very high-fidelity images that are comparable to the ones generated by SPADE. In some cases, we find our method is even slightly better than SPADE. In contrast, because of semantic information lost problem existing in common normalization layers, the results generated by Pix2pixHD are worse than both SPADE and our CLADE. In \Fref{fig:results_coco}, some visual results on the \textit{COCO-Stuff} dataset are provided. Compared to \textit{ADE20k}, \textit{COCO-stuff} has more categories and contains more small objects, so it is more challenging. However, our method can still work very well and generate high-fidelity results. According to results in \Fref{fig:results_city}, a similar conclusion can also be drawn for higher-resolution semantic image synthesis on the \textit{Cityscapes} dataset ($256\times512$).

\begin{figure*}[tp]
  \centering
  \includegraphics[width=\linewidth]{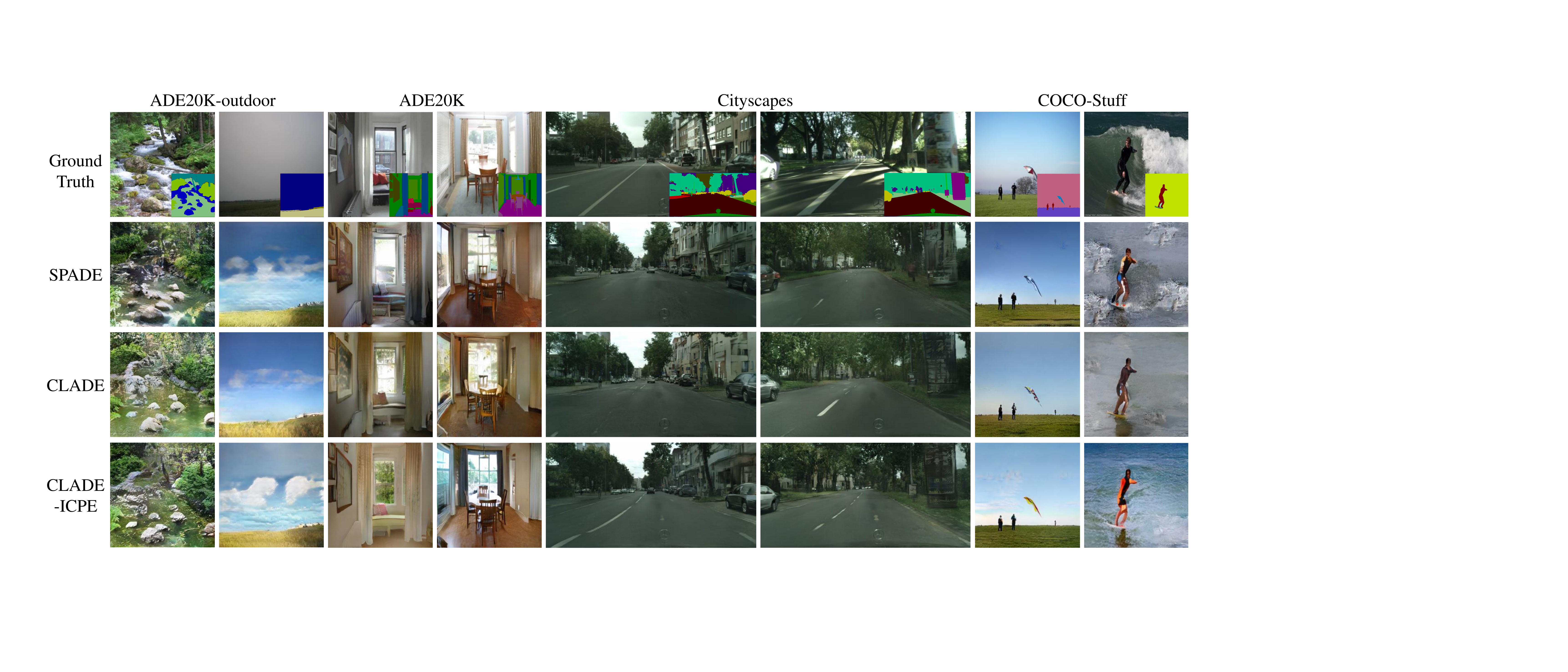}
  \caption{Visual comparison results on \textit{ADE20k-outdoor}, \textit{ADE20k}, \textit{Cityscapes} and \textit{COCO-Stuff} datasets with or without position prior. We also show the results of SPADE as a reference.}
  \label{fig:cop_dist}
\end{figure*}

We also show the results in \Fref{fig:cop_dist} to compare the visual effect of intra-class spatial-adaptiveness. Given additional spatial information, we can see richer details from the results. Taking \textit{ADE20k} dataset as an example, SPADE and CLADE can only give a blurred view out of the window, while CLADE-ICPE can generate a high-quality view with rich textures. In particular, for some classes with large regions, both SPADE and CLADE produce repeated or blurry pattern (see the last column of \Fref{fig:cop_dist}) because they cannot differentiate the difference between different positions within the same category. In contrast, CLADE-ICPE can produce vivid textures with the spatial guidance of the positional encoding map.

\begin{figure}
    \centering
    \includegraphics[width=\linewidth]{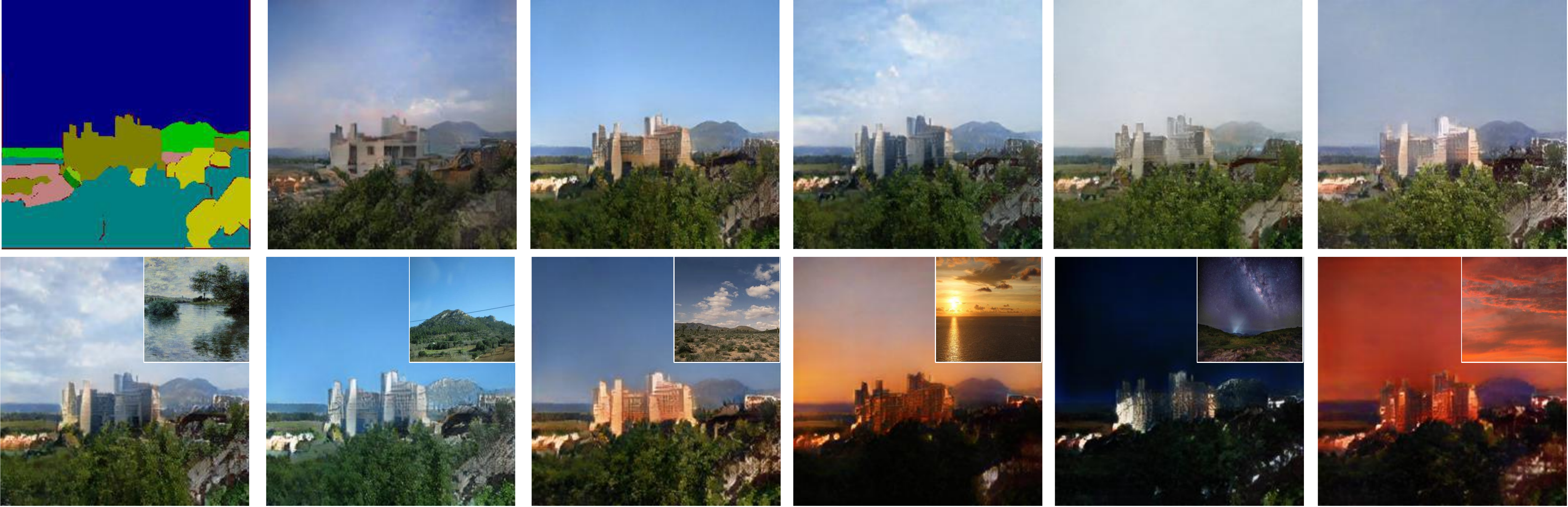}
    \caption{Multi-modal semantic synthesis results guided by different noise vectors (top row) or reference style images (bottom row). Obviously, our method can produce very diverse realistic images.}
    \label{fig:multi_modal}
\end{figure}

\begin{table*}[tp]
    \centering
    \caption{Ablation results on \textit{ADE20k-outdoor} and \textit{Cityscapes} by mixing SPADE and CLADE with the transition points at different resolutions. Here C and S represent the CLADE and SPADE layers respectively. The values in parentheses indicate the numbers of ResBlks use the specified normalized layer.}
    \label{tab:my_label}
    \setlength{\tabcolsep}{1.8mm}
    \begin{tabular}{c|c|c|c|c|c|c|c|c}
    \hline
    \multirow{3}{*}{ADE20k-outdoor} & Method & C(1-7) & S(1)+C(2-7) & S(1-2)+C(3-7) & S(1-4)+C(5-7) & S(1-5)+C(6-7) & S(1-6)+C(7) & S(1-7) \\
    \cline{2-9}
    & mIoU $\uparrow$ & 18.71 & 19.28 & 18.48 & 19.06 & \textbf{19.68} & 19.63 & 19.30 \\
    \cline{2-9}
    & Runtime (s) $\downarrow$ & \textbf{0.024} & 0.025 & 0.025 & 0.028 & 0.029 & 0.033 & 0.042 \\
    \hline
    \multirow{3}{*}{Cityscapes} & Method & C(1-7) & S(1)+C(2-7) & S(1-2)+C(3-7) & S(1-4)+C(5-7) & S(1-5)+C(6-7) & S(1-6)+C(7) & S(1-7)\\
    \cline{2-9}
    & mIoU $\uparrow$ & 60.44 & 61.25 & \textbf{62.14} & 62.08 & 62.00 & 61.47 & 61.95\\
    \cline{2-9}
    & Runtime (s) $\downarrow$ & \textbf{0.039} & 0.040 & 0.040 & 0.043 & 0.048 & 0.057 & 0.065 \\
    \hline
    \end{tabular}
    \label{tab:mix}
\end{table*}

\subsection{Multi-Modal and Style-Guided Synthesis}
As mentioned above, it is easy for our method to support multi-model and style-guided synthesis by introducing an extra style encoder before the generator network. Specifically, we get different style vectors either by random sampling or feeding different reference images into the style encoder, and then input these style vectors into the generator network to produce diverse images. In \Fref{fig:multi_modal}, some visual results are shown. The results in the top row demonstrate that our method can synthesize diverse images with the same semantic layout. Similarly, as shown in the bottom row, different reference images can be used to further control the global style of the generated images, including but are not limited to sunny days, dusk, night, etc.

\subsection{Ablation Analysis of Combining CLADE and SPADE} 
In the method part, we have shown that, for higher resolution layers, the distributions of $\gamma,\beta$ in SPADE are more centralized (\Fref{fig:feature}) and the corresponding extra computation cost is also more significant (\Fref{fig:ratio}). And for low resolution layers,  $\gamma,\beta$ is less centralized and can supply some spatial variance. In contrast, the basic CLADE is only class-adaptive but not spatial-adaptive. Therefore, it is intuitive to use SPADE on lower-resolution layers and CLADE on higher-resolution layers to achieve better balance between generation quality and efficiency. To verify this point, we mix SPADE and CLADE with the transition points at different resolution layers, and test the performance on the \textit{ADE20k-outdoor} and \textit{Cityscapes} datasets. In the original SPADE generator, there are seven SPADE ResBlks which are numbered from 1 to 7 as the resolution increases. The second and third ResBlks are at the same resolution if the resolution of the synthesized image is $256\times 256$, otherwise they are at different resolutions.  

As shown in \Tref{tab:mix}, the average running time decreases with the use of more CLADE layers, which is in line with our expectations. More interestingly, by using SPADE on low-res layers and CLADE starting from the middle ResBlks (e.g. 6th and 7th on ADE20k-outdoor dataset,  and 3rd, 5th and 6th on Cityscapes dataset), it can even achieve slightly higher mIoU than using SPADE on all layers while being more efficient. 

\subsection{Ablation Analysis of Intra-Class Positional Encoding}
Although the introduction of the positional encoding can provide prior spatial information within the semantic category and help synthesize richer details, how to properly utilize this information is not trivial. Empirically, we find that inappropriate use may even be harmful. Here we study three different ways to apply the positional encoding map:
\begin{itemize}
    \item Similar to pix2pixHD, the positional encoding map is directly concatenated with the downsampled semantic mask as extra channels and fed to the generator. This version is called \textbf{+disti}.
    \item The positional encoding map is first transformed with one $1\times1$ convolutional layer and then concatenated with each normalized features (after CLADE layer) as extra channels. In other words, the positional encoding is embedded into each normalized features.
    This version is called \textbf{+distf}.
    \item Following the design of SPADE that modulates features with spatially-adaptive parameters, as described in \Sref{sec:ICPE}, the positional encoding map is used to modulate the original semantic-adaptive modulation parameters of CLADE. This version is called \textbf{+distp}.
\end{itemize}

In the \Tref{tab:cop_dist}, we compare these three variants with the original CLADE, in terms of FID on the \textit{ADE20k}, \textit{ADE20k-outdoor} and \textit{Cityscapes} datasets. It can be seen that,  \textbf{+distp} achieves the best performance of FID on these datasets, while \textbf{+disti} is the worst. Specifically, by comparing \textbf{+distf} and \textbf{+disti}, we can easily observe that adding the spatial information at different feature levels is beneficial. And by comparing \textbf{+distf} and \textbf{+distp}, we can find that the concatenation of the positional encoding feature with normalized features along the channel dimension is not as effective as the element-wise multiplication used by \textbf{+distp}. 

Particularly, the performance gain on the Cityscapes dataset is much more significant than that on the ADE20k dataset. We think it should be because \textit{Cityscapes} contains many large-area categories with clear internal structure, such as buildings and cars. By comparison, though \textit{ADE20k-outdoor} also has some large-area categories like sky and sea, they have relatively less complex internal structures, thus benefitting less from spatial adaptiveness. 

\begin{table}[tp]
    \centering
    \footnotesize
    \caption{Comparison with different positional encoding map embedding on the \textit{ADE20k}, \textit{ADE20k-outdoor} and \textit{Cityscapes} datasets in terms of FID. Baseline denotes the original CLADE without position prior, \textbf{+distp} is the version called CLADE-ICPE in \Tref{tab:performance}.}
    \begin{tabular}{c|c|c|c|c}
        \hline
        \diagbox{Dataset}{Method} & baseline & +disti & +distf & +distp \\
         \hline
         ADE20k & 30.48 & 31.75 & 31.13 & \textbf{28.69}\\
         \hline
         ADE20k-outdoor & 46.37 & 48.67 & 46.81 & \textbf{45.59}\\
         \hline
         Cityscapes & 50.62 & 50.50 & 48.07 & \textbf{42.39} \\
        \hline
    \end{tabular}
    \label{tab:cop_dist}
\end{table}

\begin{table*}[t]
    \centering
    \caption{Performance and complexity comparison when applying CLADE onto some recent SPADE-based methods. All the models are trained with the same settings by using the official code. }
    \small
    \setlength{\tabcolsep}{1.8mm}
    \begin{tabular}{c|c|c|c|c|c|c|c}
    \hline
        Dataset & Method & mIoU $\uparrow$ & accu $\uparrow$ & FID $\downarrow$ & Params (M) $\downarrow$ & FLOPs (G) $\downarrow$ & Runtime (s) $\downarrow$\\
        \hline
        \multirow{3}{*}{CelebAMask-HQ} & SEAN & \textbf{75.94} & 95.03 & 24.30 & 266.9 & 420.8 & 0.165\\
         & SEAN-CLADE & 74.83 & \textbf{94.51} & \textbf{20.35} & \textbf{241.3} & \textbf{247.1} & \textbf{0.152}\\
         \cline{2-8}
         & GroupDNet & 76.13 & 95.21 & 29.39 & 145.3 & 225.5 & 0.090\\
         & GroupDNet-CLADE & \textbf{76.70} & \textbf{95.38} & \textbf{29.30} & \textbf{134.6} & \textbf{213.6} & \textbf{0.074}\\
        \hline
        \multirow{3}{*}{DeepFashion} & SEAN & 76.28 & 97.46 & 7.37 & 223.2 & 342.9 & 0.165\\
         & SEAN-CLADE & \textbf{76.32} & \textbf{97.52} & \textbf{7.33} & \textbf{197.8} & \textbf{247.1} & \textbf{0.152}\\
         \cline{2-8}
         & GroupDNet & 76.19 & 97.48 & \textbf{9.72} & 96.3 & 291.6 & 0.062\\
         & GroupDNet-CLADE & \textbf{76.82} & \textbf{97.67} & 9.79 & \textbf{79.2} & \textbf{118.5} & \textbf{0.042}\\
        \hline
        \multirow{3}{*}{Cityscapes} & SEAN & 59.02 & \textbf{93.21} & 53.85 & 330.4 & 681.8 & 0.507\\
         & SEAN-CLADE & \textbf{60.11} & 93.15 & \textbf{52.76} & \textbf{304.5} & \textbf{476.1} & \textbf{0.471}\\
         \cline{2-8}
         & GroupDNet & 59.20 & 92.78 & \textbf{41.12} & 76.5 & 463.6 & 0.224\\
         & GroupDNet-CLADE & \textbf{59.82} & \textbf{92.83} & 42.10 & \textbf{57.7} & \textbf{434.6} & \textbf{0.128}\\
        \hline
    \end{tabular}
    \label{tab:extend}
\end{table*}{}

\subsection{Generalization Ability to SPADE-Based Methods}
To demonstrate the general applicability, we further replace the SPADE layer with the proposed CLADE layer for some recent SPADE-based methods and show the results in \Tref{tab:extend}. Without loss of generality, we still focus on the semantic image synthesis task and select two representative methods: GroupDNet~\cite{zhu2020semantically} and SEAN~\cite{zhu2020sean}. GroupDNet is a semantic-level multimodal image synthesis method that achieves great success on the \textit{DeepFashion} dataset, while SEAN focuses on face image synthesis and shows excellent performance on the \textit{CelebAMask-HQ} dataset. Therefore, considering the performance of these two methods on their respective datasets, we also choose to conduct experiments on these two datasets, and add \textit{Cityscapes} dataset as a supplement. All the models are trained with the same settings in the official codes and the only difference is the normalization layer. 

The detailed comparison results are shown in  the \Tref{tab:extend}. In general, after replacing SPADE with CLADE, the original performance of such methods are almost not affected but the parameter number and computational overhead are significantly reduced. More interestingly, most of the metric indicators (including mIoU, accu and FID) are even improved slightly.
In details, for SEAN, the performance on the \textit{CelebAMask-HQ} dataset in terms of FID is significantly improved. As for model parameters, a reduction in model size of about 20M on different datasets is observed, which is consistent with the comparison between SPADE and CLADE in \Tref{tab:performance}. Similarly, for different data sets, the running time is correspondingly reduced by $7\%$ to $40\%$. It seems that GroupDNet has fewer parameters than SPADE on \textit{Cityscapes} dataset, but its efficiency is still not satisfactory in terms of the FLOPs and running time. In contrast, CLADE, as an efficient counterpart, can be easily applied to the SPADE-based methods or other methods which use semantic masks as input. 

\section{Conclusion}
In this paper, we conduct an in-depth analysis on the spatially-adaptive normalization used in semantic image synthesis. We observe that its most essential advantage comes from semantic-awareness instead of spatial-adaptiveness as originally suggested in~\cite{park2019semantic}. Motivated by this observation, we design a more efficient conditional normalization structure CLADE. Compared to SPADE, CLADE can achieve comparable synthesis results but greatly reduce the parameter and computation overhead. To introduce true spatial adaptiveness, we further explore the role of position prior and propose an improved version of CLADE by modulating the parameters of CLADE through an extra intra-class positional encoding. We further adopt CLADE in some recent SPADE-based methods and get comparable or even better results with greatly reduced parameters and computational costs. 

\vspace{1em}
\noindent\textbf{Acknowledgement.} This work is supported by the National Natural Science Foundation of
China (Grant No. U20B2047). 
Jing Liao is partially supported by an ECS grant from the Research Grants Council of the Hong Kong (Project No. CityU 21209119) and an APRC grant from CityU, Hong Kong (Project No. 9610488). Gang Hua is partially supported by National Key R\&D Program of China Grant 2018AAA0101400 and NSFC Grant 61629301. 


%





\ifCLASSOPTIONcaptionsoff
  \newpage
\fi



%

\bibliographystyle{IEEEtran}
\bibliography{egbib}



%
\end{document}